\documentclass{nature}
\usepackage{graphicx}
\usepackage{sectsty}
\usepackage{amssymb} 
\usepackage{amsmath} 

\usepackage{color}
\definecolor{gray}{rgb}{0.8,0.8,0.8}
\definecolor{cyan}{cmyk}{1,0,0,0}
\definecolor{blue}{rgb}{0,0,0.7}
\definecolor{red}{rgb}{0.5,0.0,0}
\definecolor{darkgreen}{rgb}{0,0.6,0}
\definecolor{darkergreen}{rgb}{0,0.3,0}
\definecolor{orange}{rgb}{1,0.5,0}
\definecolor{magenta}{cmyk}{0,1,0,0}
\definecolor{darkyellow}{cmyk}{0,0,0.5,0}
\definecolor{purple}{cmyk}{0.5,1,0,0}

\graphicspath{{Image/}}

\title{Virtual Wave Optics for Non-Line-of-Sight Imaging}

\author{Xiaochun Liu$^{1}$, Ib\'{o}n Guill\'{e}n$^{2}$, Marco La Manna$^{3}$, Ji Hyun Nam$^{1}$, Syed Azer Reza$^{3}$, Toan Huu Le$^{1}$, Diego Gutierrez$^{2}$, Adrian Jarabo$^{2}$, Andreas Velten$^{1,3}$}

\begin{document}
\maketitle

\begin{affiliations}
\item Department of Electrical and Computer Engineering, University of Wisconsin Madison
\item Graphics and Imaging Lab, Universidad de Zaragoza - I3A
\item Department of Biostatistics and Medical Informatics, University of Wisconsin Madison
\end{affiliations}

\begin{abstract}
\end{abstract}

\newcommand{\px}[1]{\mathbf{#1}}

\textit{\textbf{The work related to this paper is online in Nature, title "Non-line-of-sight imaging using phasor-field virtual wave optics", link: https://www.nature.com/articles/s41586-019-1461-3.}}

\textbf{Non-Line-of-Sight (NLOS) imaging allows to observe objects partially or fully occluded from direct view, by analyzing indirect diffuse reflections off a secondary, relay surface. Despite its many potential applications\cite{Kirmani2011,Gupta2012,Velten2012,Katz2012,Heide2014,Laurenzis2014,Buttafava2015,Arellano2017,OToole2018}, existing methods lack practical usability due to several shared limitations, including the assumption of single scattering only, lack of occlusions, and Lambertian reflectance. We lift these limitations by transforming the NLOS problem into a virtual Line-Of-Sight (LOS) one. Since imaging information cannot be recovered from the irradiance arriving at the relay surface, we introduce the concept of the \textit{phasor field}, a mathematical construct representing a fast variation in irradiance. We show that NLOS light transport can be modeled as the propagation of a phasor field wave, which can be solved accurately by the Rayleigh-Sommerfeld diffraction integral. We demonstrate for the first time NLOS reconstruction of complex scenes with strong multiply scattered and ambient light, arbitrary materials, large depth range, and occlusions. Our method handles these challenging cases without explicitly developing a light transport model. By leveraging existing fast algorithms\cite{Arellano2017}, we outperform existing methods in terms of execution speed, computational complexity, and memory use. We believe that our approach will help unlock the potential of NLOS imaging, and the development of novel applications not restricted to lab conditions. For example, we demonstrate both refocusing and transient NLOS videos of real-world, complex scenes with large depth.
}

\begin{figure}[ht!]
\includegraphics[width = \columnwidth]{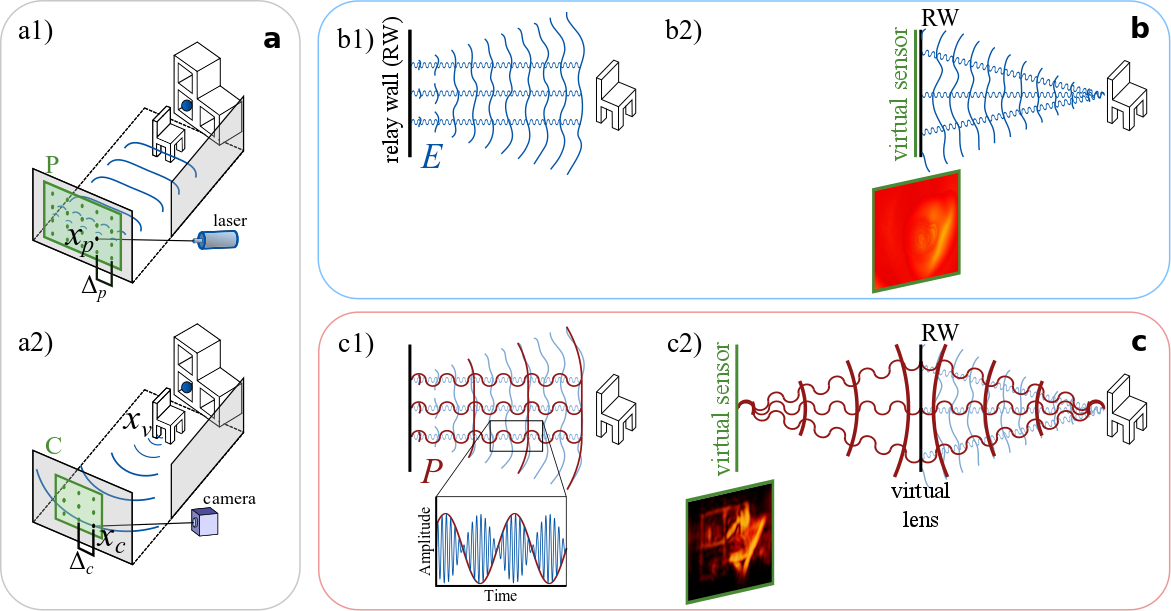}
\caption{\textbf{NLOS as a virtual LOS imaging system.}
\textbf{a)} Classic setup: A pulsed laser sequentially scans a relay wall (a.1); the light reflected back from the scene onto the wall is recorded at the sensor (a.2).
\textbf{b)} The superposition of the signal from all the scanned points results in a planar electromagnetic field $E$ (b.1). After interacting with the scene, a spherical wavefront is reflected back. Phase information is lost at the diffuse relay wall, hampering the imaging process (b.2). 
Existing NLOS techniques like filtered backprojection\cite{Arellano2017} or deconvolution\cite{OToole2018} aim to reproduce a sharper result from this suboptimal information.
\textbf{c)} In our approach, we instead propagate a phasor field $\mathcal{P}$, representing the complex envelope of the electromagnetic field $E$ (c.1). The phase of $\mathcal{P}$ remains known and controllable at the relay wall, which now acts as a \textit{virtual lens} from which the image can be brought back to focus on a virtual sensor placed away from it (c.2). This approach effectively creates a virtual LOS system, thus bringing all existing LOS imaging techniques into the NLOS realm.
}
\label{fig:overview}
\end{figure}

We have recently witnessed large advances in transient imaging techniques\cite{Jarabo2017}, employing streak cameras\cite{Velten2013,Heshmat2014}, gated sensors\cite{Laurenzis2014}, amplitude-modulated continuous waves\cite{Kirmani2013,Gupta2015,Tadano2016}, or interferometry\cite{Gkioulekas2015}. Access to time-resolved image information has in turn led to advances in imaging of objects partially or fully hidden from direct view\cite{Velten2012,Kirmani2011,Katz2012,Heide2014,Gupta2012,Buttafava2015,Katz2014,Wu2018}. 
Figure~\ref{fig:overview}.a shows the basic configuration of a NLOS system, where light bounces off a relay wall, travels to the hidden scene, then goes back to the relay wall and then finally the sensor. 

Recent NLOS reconstruction methods can compute exact inverse operators of forward light transport models\cite{Velten2012,Gupta2012,Buttafava2015,Heide2014,OToole2018}. However, they assume a discrete ray-based light propagation and do not take into account multiply scattering, occlusions and clutter, or surfaces with anisotropic reflectance. Moreover, the depth range that can be recovered is also limited, due partially to the difference in intensity between first- and higher-order reflections. 
As such, existing methods are limited to carefully controlled cases, imaging isolated objects of simple geometry. 

On the other hand, LOS imaging systems across all size and resolution scales provide methods to image complex scenes including extreme amounts of MPI, large depth range, and varying scene albedos. We show that the data collected by any band-limited, NLOS imaging system is fundamentally equivalent to the data collected by a conventional (virtual) LOS system, placed at the diffuse relay wall towards the hidden scene, and can be described by the Rayleigh-Sommerfield diffraction (RSD) integral\cite{Sommerfeld1954,Born1999}.
However, when an optical wavefront interacts with a diffuse surface, phase information is lost, destroying valuable imaging information (Figure~\ref{fig:overview}.b). 

Instead, we introduce the \textit{virtual phasor field} (Figure~\ref{fig:overview}.c), a mathematical construct which represents the  the light intensity deviation at a point in space from its long time average. 
%
Phasor formalism has been used together with Fourier domain ranging in LOS imaging techniques\cite{Kadambi2016,Gupta2015} to describe the modulated light signal emitted by specific imaging hardware. We show that a similar description can be used to model the physics of light transport through the scene, which allows us to formulate any NLOS imaging problem as a wave imaging problem. 
Since all wave-based imaging systems (optical, LiDAR, ultrasound...) are based on the same mathematical principles of wave propagation, this allows us to derive a class of methods that broadly links all NLOS imaging systems to LOS wave imaging systems,
and to transfer well-established insights and techniques from classical optics into the NLOS domain.
%
 
The RSD integral describes the propagation of a monochromatic wave $P_\omega(x,t)$ between a source plane $S$ and a destination plane $D$ as  

\begin{equation}
\label{eq:huygens}
\mathcal{P}_\omega({x}_d,t) =\gamma \int_{S} {\mathcal{P}_\omega({x}_s,t)\frac{e^{ik|{x}_d-{x}_s|}}{|{x}_d-{x}_s|}d{x}_s},
\end{equation}
%
%
where $x_s$ and $x_d$ are points on $S$ and $D$ respectively, $\gamma$ is a constant equal to $1/i\lambda$, where $\lambda$ represents wavelength, $\omega=c/\lambda$ is the frequency of the wave, $c$ its propagation speed, and $k = 2\pi/\lambda$ is the wave number. 
%
 
%
A wave-based imaging system captures an incoming wavefront at points $x_d$ on its aperture as a function of time $t$, and transforms it into an image of the scene as $I(t) = \Phi\left(\mathcal{P}_\omega(x_d, t)\right)$, where the operator $\Phi$ denotes the image formation process, which depends on the particular imaging system. The nature of $\mathcal{P}_\omega(x,t)$ depends on the particular imaging method under consideration, e.g., acoustic pressure for ultrasound, electromagnetic field for optical and microwave imaging, etc. In our work $\mathcal{P}_\omega(x,t)$ represents our virtual phasor field, 
%
which can be computed from the electromagnetic field $E$ by
\begin{equation}
 \label{eq:P_field_def}
\mathcal{P}_\omega(x,t) = \int_{t-\tau/2}^{t+\tau/2}{|E(x,t')|^2\,dt'}-\frac{1}{T}\int_{t-T/2}^{t+T/2}{|E(x, t')|^2\,dt'},
\end{equation}
where $|.|$ indicates absolute value (intensity magnitude for irradiance $E$), $T$ represents a long term average intensity of the signal as would be seen by a conventional photo-detector or camera pixel, and $\tau<<T$. 
In the presence of a rough aperture such as diffuse wall, phase fluctuations in the phasor field are negligible compared to the wavelength of the phasor field wave\cite{Reza2018}. In the supplemental material (section \ref{supp:RSD}), we show that light propagating from a diffuse surface $S$ can be described in terms of our phasor field by means of the RSD (Equation~\ref{eq:huygens}). 

Multiply scattered light in the hidden scene can be sequentially described by the RSD operator. 
Since this process is linear in space and time invariant\cite{Sen2005,OToole2014}, we can characterize light transport through the scene as a linear transfer function $H(x_p \rightarrow x_c, t)$, where $x_p$ represents a point in the \textit{virtual projector} aperture $P$ emitting light, and $x_c$ a point in the \textit{virtual camera} aperture $C$ capturing it (see the supplementary information \ref{supp:RSD_linearity} for a detailed derivation).
Any imaging system can be posed as an inverse wave propagation operator attempting to reconstruct properties of the scene from $H$, which can be measured in a NLOS imaging system by sequentially illuminating the relay wall with short $\delta$ pulses from a femtolaser. 
%

We first determine the outgoing illumination wavefront $\mathcal{P}_\omega(x_p, t)$ used in the LOS imaging system we seek to emulate; methods to determine the wavefront of known illumination systems are available in the literature\cite{Goodman2005}. 
%
We then compute the field $\mathcal{P}_\omega$ at the camera aperture $C$ using the measured scene response function $H$ as
\begin{equation}
\label{eq:projector}
\mathcal{P}_\omega(x_c,t) =  \int_P \mathcal{P}_\omega(x_p,t) \star H(x_p \rightarrow x_c, t) dx_p,
\end{equation}
where $\star$ represents the convolution operator in the temporal domain.
Finally, the image is obtained as $I = \Phi\left(\mathcal{P}_\omega(x_c, t)\right)$, revealing the hidden scene encoded in $H$.
In our virtual LOS system, the wavefront $\mathcal{P}_\omega(x_p, t)$ and the imaging operator $\Phi$ are implemented computationally, thus despite being based on physical LOS systems\cite{Goodman2005} they are not bounded by the limitations imposed by actual hardware. In this work we apply this method by implementing three virtual imaging systems based on three common LOS imaging systems. A conventional photography camera capable of capturing 2D NLOS imaging without knowledge of the timing or location of the illumination source, a femto photography system capable of capturing transient videos of the hidden scene, and a confocal imaging system robust to MPI. An in-depth description of the example imaging systems  provided in the supplemental material \ref{supp:projectorcamera}.
\begin{figure*}[ht!]
\centering
\resizebox{\textwidth}{!}{%
\includegraphics[height=5cm]{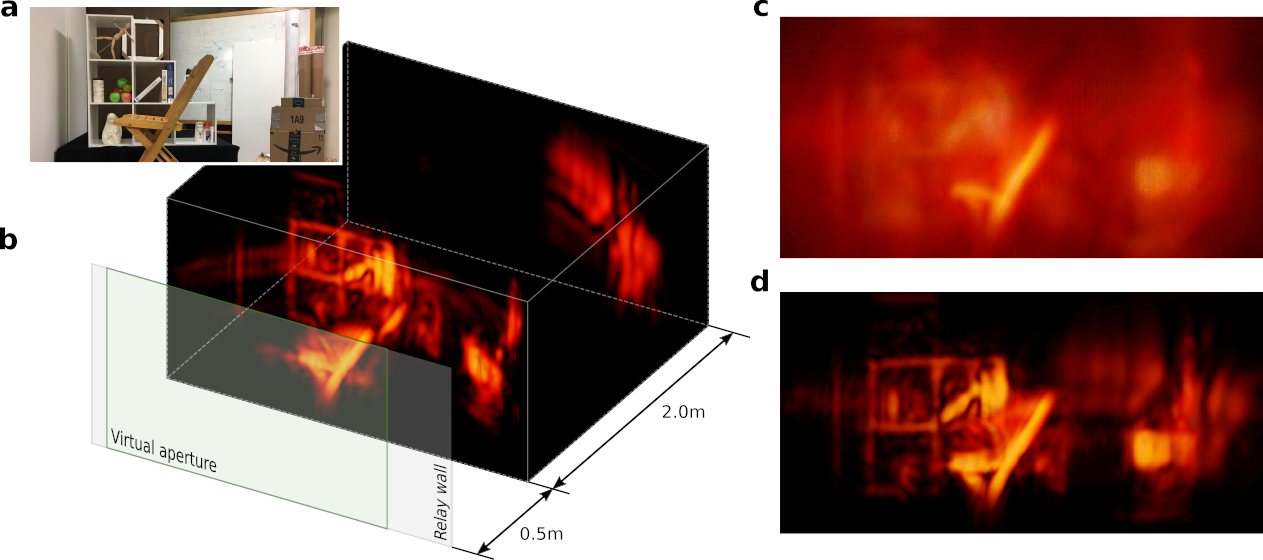}
}
\caption{\textbf{Reconstructions of a hidden, complex real scene.}
\textbf{a} Photograph of the scene as seen from the relay wall. The scene contains occluding geometries, multiple anisotropic surface reflectances, large depth, and strong ambient and multiply scattered light.
\textbf{b} 3D visualization of the proposed reconstruction using phasor fields, revealing the geometry and spatial distribution of the different objects in the scene, as well as its depth range. We include the relay wall location and the coverage of the virtual aperture for illustration purposes.
\textbf{c} Current techniques fail to reconstruct the scene.
\textbf{d} Frontal view of the proposed reconstruction using phasor fields. 
}
\label{fig:main_results}
\end{figure*}

Our capture system consists on a Onefive Katana HP laser (1~W at 532~nm, and a pulse width of about 35 ps used at a repetition rate of 10 MHz), and a gated SPAD detector processed by a time-correlated single photon counter (TCSPC, PicoQuandt HydraHarp) with a time resolution of about 30~ps and a dead time of 100 ns.
%
The actual time resolution of our system is approximately 65~ps, a combination of the pulse width of the laser, and the jitter of the system. 
The spatial resolution is $\Delta_x=0.61 d\lambda/L$, where $d$ is the aperture diameter, and $L$ is the imaging distance.
%
The distance $\Delta_p$ between sample points $x_p$ in $P$ (see Figure \ref{fig:overview}) has to be small enough to sample $H$ at the phasor field wavelength, which we fix at $\Delta_p=\lambda/4$.
Our prototype implementation uses a single SPAD sensor, which suffices to reconstruct complex hidden scenes. We plan to replace it with an array of sensors to speed up capture times. 
Our algorithm is bounded by the cost of backprojection during reconstruction, which we run on the GPU; this allows us to limit its cost to $O(N^3)$, as efficient as state-of-the-art backprojection\cite{Arellano2017}, and faster than the recent confocal approach\cite{OToole2018}. A comprehensive survey of the cost of the different NLOS techniques can be found in the supplementary Section~\ref{supp:complexity}.

\begin{figure*}[ht!]
\centering
\resizebox{\textwidth}{!}{%
\includegraphics{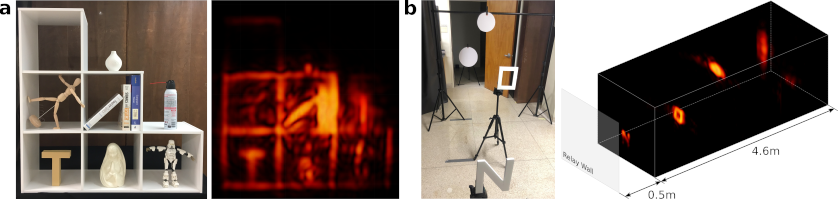}%
}
\caption{\textbf{Robustness of our technique}. \textbf{a} Reconstruction in the presence of strong ambient illumination (all the lights on during capture). \textbf{b} Hidden scene with a large depth range, leading to very weak signals from objects farther away.
}
\label{fig:additional_results}
\end{figure*}

\begin{figure*}[ht!]
\centering
\includegraphics[width=\textwidth]{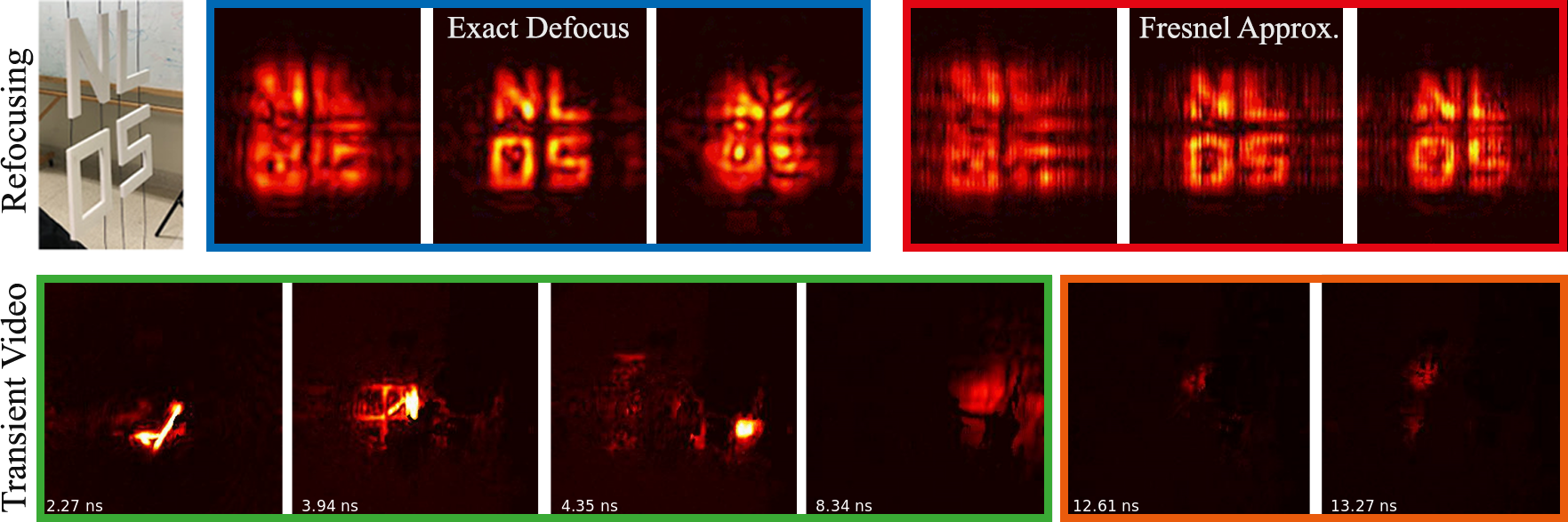}%
\caption{\textbf{Additional novel NLOS imaging applications of our method.} \textbf{(Top)} NLOS refocusing: The hidden letters (left) are progressively brought in and out of focus as seen from a virtual camera at the relay wall, using the exact lens integral (middle, in blue) and the Fresnel approximation (right, in red). 
Each set of images shows the scene focused at different depths. The reconstruction is performed without knowledge of the locations and timing of the laser illumination used in the capture process. \textbf{(Bottom)} NLOS femto-photography: Example frames of light traveling through the hidden office scene shown in Figure~\ref{fig:main_results} when illuminated by a pulsed laser. Numbers indicate the propagation time from the relay wall. The first four frames (green) show how light first illuminates the chair, then it propagates to the shelf, then the boxes far on the right, and finally the back wall three meters away. Moreover, our technique is able to reconstruct higher-order scattering beyond the first interaction. The last two frames (in orange) show how the pulse of light travels from the wall back to the scene. Refer to the supplementary video for the full animations.
}
\label{fig:additional_imaging}
\end{figure*}

We can describe novel NLOS imaging systems and applications, leveraging the wealth of tools and processing methods available in LOS imaging. Figure~\ref{fig:additional_imaging} (top) demonstrates NLOS refocusing an image of our virtual photography camera, both exact and using a faster Fresnel approximation, while the bottom line shows NLOS femtophotography. A description of the Fresnel approximation as well as the LOS projector-camera functions used in these examples appear in the supplemental (Sections~\ref{supp:fresnel} and ~\ref{supp:pcfunctions}).

Figure~\ref{fig:main_results} shows the result of reconstructing a complex hidden scene with the virtual confocal camera. The scene contains multiple objects with occlusions distributed over a large depth, a wide range of surface reflectances and albedos, and strong interreflections. For comparison purposes, we show also the result of current techniques, including a Laplacian filter, which fail to reconstruct most of the details (refer to Figure~\ref{fig:overview}, b.2). We provide additional comparisons with previous techniques in the supplemental Section~\ref{supp:CNLOS_cmp}, making use of a public dataset\cite{OToole2018}.

In Figure~\ref{fig:additional_results} we demonstrate the robustness of our method when dealing with challenging scenarios, including strong multiply scattering and ambient illumination from ambient light sources (3.a), or a high dynamic range from objects spanning large depths (3.b). 

These scenes are significantly more complex than any NLOS reconstruction shown so far in the literature, and highlight the primary benefit of our approach: By turning NLOS into a virtual LOS system, the intrinsic limitations of previous approaches no longer apply. 

By bringing NLOS into the real of virtual LOS, our work enables a new class of NLOS imaging methods that leverage existing wave-based imaging methods. It provides superior robustness in the presence of noise, ambient illumination and interreflections, anisotropic reflectance, and overall scene complexity, leading to higher reconstruction quality with respect to previous methods. 
In addition, our approach is fast, memory-efficient, and lacks computational complexity since it does not require inverting a light transport model.
Working in a virtual LOS domain, we foresee other advances in the near future, such as looking around a \textit{second} corner, or selection and manipulation of individual light paths to isolate specific aspects of the light transport in different NLOS scenes.

\subsection*{Acknowledgements}

This work was funded by DARPA through the DARPA REVEAL project, the NASA NIAC program, the AFOSR Young Investigator Program, the European Research Council (ERC) under the EU's Horizon 2020 research and innovation programme (project CHAMELEON, grant No 682080), the Spanish Ministerio de Econom\'ia y Competitividad (project TIN2016-78753-P), and the BBVA Foundation (Leonardo Grant for Researchers and Cultural Creators). We would like to acknowledge Jeremy Teichman for hepful insights and discussions in developing the phasor field model. We would also like to acknowledge Mauro Buttafava, Alberto Tosi, and Atul Ingle for help with the gated SPAD setup.

\newpage
\appendix
\section{Supplemental material}

\subsection{The phasor field Rayleigh-Sommerfeld integral}
\label{supp:RSD}

Here we derive the Rayleigh-Sommerfeld diffraction integral for the phasor field. Consider a point light source at a location $x_s$ that emits light with a sinusoidal time varying intensity $L(t)=L_0\,(e^{i\omega t}+1)$ with amplitude $L_0$ and modulation frequency $\omega$. More formally, $L(t)$ and $L_0$ relate with the electromagnetic field $E(x,t)$ as $L(x,t)=\int_{t-\tau/2}^{t+\tau/2}|E(x,t')|^2 dt'$ and $L_0(x) = \int_{-\infty}^{+\infty}|E(x,t')|^2 dt'$, with $\tau$ a sufficiently small value. 
We define the phasor field $\mathcal{P}(x,t)$ at a point in space as 
%
\begin{equation}
\mathcal{P}(x,t)=L(x,t)-L_0(x).
\end{equation}
Let us now consider the phasor field wave of a monochromatic point light source at location $x_{s}$ oscillating at at a frequency $\omega$,
\begin{equation}
\mathcal{P}_\omega(x_{s},t)=\mathcal{P}_0(x_s)e^{i\omega t}.
\end{equation}
To determine the light intensity and thereby the phasor field at any point in space and time $(x_d, t)$ we have to account for the travel time from $x_s$ to $x_d$ defined as $t_p=|x_d-x_s|/c$, with $c$ the propagation speed, and the radial drop off in light intensity,
\begin{equation}
\mathcal{P}_\omega(x_d, t) =
	\mathcal{P}_0 \frac{e^{i\omega (t + t_p)}}{|x_d-x_s|^2} =
    \mathcal{P}_0 \frac{e^{i\omega (t + |x_d-x_s|/c)}}{|x_d-x_s|^2} =
    \mathcal{P}_0 \frac{e^{i\omega t + ik|x_d-x_s|}}{|x_d-x_s|^2}
\end{equation}
where $k$ is the wave number. If instead of a single light source we have a collection of incoherent sources comprising a surface $S$, we have
\begin{equation}
\mathcal{P}_\omega(x_d, t)=\int_S{\mathcal{P}_0 \frac{e^{i\omega t + i k|x_s-x_d|}}{|x_s-x_d|^2} dx_s}.
\end{equation}
This equation looks like the Rayleigh-Sommerfeld propagator, except for the squared denominator. We approximate $|x_s-x_d|^2\approx |x_s-x_d||<\bar S>-x_d|$ where $<\bar S>$ is the average position of all source points. Pulling this constant term out of the integral, we obtain
\begin{align}
\label{eq:PhasorRSD}
\mathcal{P}_\omega(x_d, t)& \approx
	\frac{1}{|<\bar S>-x_d|}\int_S{\mathcal{P}_0 \frac{e^{i\omega t + i k|x_s-x_d|}}{| x_s-x_d|} dx_s} \nonumber\\ &=
    \frac{1}{|<\bar S>-x_d|}\int_S{\mathcal{P}_\omega(x_s,t) \frac{e^{i k|x_s-x_d|}}{| x_s-x_d|} dx_s}
\end{align}
which is the RSD (Equation~\eqref{eq:huygens}) for $\gamma=1 /|<\bar S>-x_d|$. This approximation does not affect the phase term, causing only a slow-varying error in amplitude. Given a known source plane, this error can be precomputed. In the interest of brevity, the propagation model is analyzed in more detail in a separate manuscript\cite{Reza2018}, which also investigates the effect of the approximation made in Equation~\ref{eq:PhasorRSD}. As we show in Section~\ref{supp:lenses}, most real imaging systems do not invert the $1/r$ term in the RSD propagator. Further research may also lead to alternative formulations of the phasor field that deal with this error in a more elegant way.

\subsection{Linearity of the RSD integral}
\label{supp:RSD_linearity}
Our objective is to show that propagation of a wave $\mathcal{P}(x_s, t)$ as defined by the RSD diffraction integral
\begin{equation}
\mathcal{P}(x_d, t) =
	\mathcal{R}(\mathcal{P}(x_s, t)) =
    \gamma \int_S{\mathcal{P}(x_s, t) \frac{e^{i k| x_d - x_s|}}{|x_d - x_s|} d x_s}
\end{equation}
is linear in the sense that for two waves $\mathcal{P}_1$ and $\mathcal{P}_2$ propagating between surfaces $S$ and $D$, we can write
\begin{equation}
\mathcal{R}(a\,\mathcal{P}_1(x_{s}, t) + b\,\mathcal{P}_2(x_{s}, t)) =
	a\,\mathcal{R}(\mathcal{P}_1(x_{s}, t)) + b\,\mathcal{R}(\mathcal{P}_2(x_{s}, t))
\end{equation}
which can be seen by writing out the integral:
\begin{equation}
\begin{split}
\mathcal{R}(\mathcal{P}(x_s, t)) =
	\gamma \int_S { [a\,\mathcal{P}_1(x_{s}, t) + b\,\mathcal{P}_2(x_{s}, t)] \frac{e^{i k| x_d - x_s|}}{|x_d - x_s|} d x_s} =\\
    a\,\gamma \int_S{\mathcal{P}_1(x_s, t) \frac{e^{i k| x_d - x_s|}}{|x_d - x_s|} d x_s} +
    b\,\gamma \int_S{\mathcal{P}_2(x_s, t) \frac{e^{i k| x_d - x_s|}}{|x_d - x_s|} d x_s} .
\end{split}
\end{equation}
From the integral it can also be seen that a convolution with a kernel in time $k(t)$ applied to the argument of $\mathcal{R}$ is equal to a convolution applied to its value,
\begin{equation}
\mathcal{R}([\mathcal{P}\star k]_t) = [\mathcal{R}(\mathcal{P})\star k]_t .
\end{equation}
Multibounce light propagation through a NLOS scene can be expressed in terms of multiple applications of the RSD operator (one for each bounce):
\begin{equation}
\mathcal{S}(\mathcal{P}(x_s, t)) =
	\mathcal{R}(\mathcal{R}(\mathcal{P}(x_s, t))) +
    \mathcal{R}(\mathcal{R}(\mathcal{R}(\mathcal{P}(x_s, t))))+...
\end{equation}

To obtain $H$ we measure the output of $\mathcal{S}$ for a set of functions $\mathcal{P}_n(x_{s,n}, t)=\delta(x_{s,n})\delta(t)$, being $\delta$ negligibly small pulses in space and time. 
Any function $\mathcal{P}(x_s, t)$ can be written as a superposition of $\mathcal{P}_n$ as
\begin{equation}
\mathcal{P}(x_s, t)=\sum_n{[\delta(x_{s,n})\delta(t) \star \mathcal{P}(x_{s,n}, t)]_t} .
\end{equation}
To compute the scene response of any arbitrary field $\mathcal{P}(x_s, t)$ we can then write:
\begin{equation}
\begin{split}
\mathcal{S}(\mathcal{P}(x_s, t))
	& = \mathcal{S}(\sum_n{[\delta(x_{s,n})\delta(t) \star \mathcal{P}(x_{s,n}, t)]_t})\\
    & = \sum_n{[\mathcal{S}(\delta(x_{s,n})\delta(t)) \star \mathcal{P}(x_{s,n}, t)]_t})\\
    & = \sum_n{[H \star \mathcal{P}(x_{s,n}, t)]_t}
    = \int_S{[H \star \mathcal{P}(x_{s}, t)]_t d x_s}
\end{split}
\end{equation}
where the integral is understood as a continuous version of the sum over points $x_{s,n}$.

\subsection{Projector and camera functions}
\label{supp:projectorcamera}

\subsubsection{Phase operator of an ideal lens}
\label{supp:lenses}
We define an ideal lens as an element that focuses a planar wavefront into a point at the focal distance $f$ from the lens. This is equivalent to converting light coming from a point $x_f$ and turning it into a planar wave, i.e. a wave with a phase $\phi$ that is independent from the position $x_l$ on the plane of the lens. Light leaving from a point at $x_f$ creates spherical wavefronts, i.e. the phase at a plane perpendicular to the z-direction at a distance $f$ from $x_f$ is
\begin{equation}
e^{i\phi}=e^{i\omega \frac{|x_f - x_l|}{c}}
\label{eq:lens}
\end{equation}
The lens phase shift has to cancel this phase term and thus the lens acts on a wavefront $\mathcal{P}_\omega(x)$ as follows
\begin{equation}
\mathcal{P}'_\omega =
	\mathcal{P}_\omega(x) \ e^{i\omega \frac{|x_f - x_l|}{c}}
\end{equation}

Consider a wave $\mathcal{P}(x, t)$ expressed as a superposition of monochromatic waves
\begin{equation}
\mathcal{P}(x, t) =
	\int_{-\infty}^{+\infty}{\mathcal{P}_\omega(x)d\omega}=
    \int_{-\infty}^{+\infty}{\mathcal{P}_0(x)e^{i\omega t}d\omega};
\end{equation}
Applying the phase shift of the lens (Eq.\ref{eq:lens}) to this wavefront yields
\begin{equation}
\label{eq:focus}
\int_{-\infty}^{+\infty}{\mathcal{P}_0(x)e^{i(\omega t-\omega\frac{|x_f - x_l|}{c})}}d\omega=\mathcal{P}(x, t-\frac{|x_f - x_l|}{c}).
\end{equation}

The phase shift of an ideal lens can thus also be described as a shift in time.

Note that the lens only cancels the phase term of the RSD propagator and completely ignores the $1/r$ falloff. Our phasor field RSD propagator differs from the physical wave RSD operator in the modeling of this falloff. Since most real imaging systems do not invert this falloff, we choose to also not correct for it. Such a correction is possible, but would add complexity to our method and may only be warranted if quantitative albedo information is to be reconstructed.

\subsubsection{Example Projector and Camera Functions}
\label{supp:pcfunctions}
Our theoretical model allows us to implement any arbitrary (virtual) camera system by defining the projector function $\mathcal{P}(x_p,t)$ and camera operator $\Phi(\mathcal{P}(x_c,t))$. Methods for modeling such function using Fourier optics are available in the literature\cite{Goodman2005}. In our work we implement three of them (see Table~\ref{table:examples}). Each has capabilities never before demonstrated in NLOS imaging:

(1) A conventional photography camera system with a monochromatic illumination source at frequency $\omega$. It reconstructs the hidden scene with low computational effort. Like a LOS photography camera it does not require knowledge of the position or timing of the light source and only requires that one or more light sources be active wile the image data is collected. In other words, reconstruction is independent on the $x_p$ argument in $H$. The projector function $\mathcal{P}(x_p, t)$ in this case can be anything. However, like in conventional imaging, the resolution of the image is determined by the temporal bandwidth of $\mathcal{P}(x_p, t)$. We thus have to choose a function with a fast temporal component or a short phasor field wavelength; $e^{i\omega t}$ for a chosen phasor field wavelength $\lambda$ fulfills this requirement. The camera operator is a lens that creates an image on a set of detector pixels that record the absolute value (actually the absolute value squared) of the field. Implementing the lens using the time shift property (Eq.~\ref{eq:focus}) yields $|\int_C{\mathcal{P}(x_c, t-\frac{1}{c}|x_v-x_c|)dx_c}|$. Note that this expression will be constant with time just like the intensity in the sensor of a LOS photography camera, so it can be evaluated at any time $t$.

The second system (2) is a femtophotography transient image capture system\cite{Velten2013}, which can capture videos of light transport through the NLOS scene. In this case we model a monochromatic point light source at a single point $x_{ls}$ which illuminates the scene with a short flash of width $\sigma$ at time $t_0$. The camera captures a frame at time $t_f$. We assume the camera focus follows the pulse such that the camera is always focused at the depth that is currently illuminated by the pulse. Like its LOS counterpart, this camera can visualize the propagation of the illumination pulse through the scene and reveal complex multibounce light transport phenomena. As a consequence this virtual camera may be used to separate light transport into direct and global components. A straight-forward implementation of this concept has two problems: Because of the large camera aperture, the depth of field of the camera is smaller than the width of the illumination pulse. This means we can see objects go in and out of focus as they are illuminated. In addition, since we have to compute backprojections for the entire scene at each frame, generating this type of video is computationally very expensive. We can solve both problems simultaneously by realizing that most voxels in the scene do not contain any object surface and thus have a zero value throughout the video. We therefore can limit ourselves to only reconstruct all time points for the voxels in the volume that have the maximum intensity along the depth or $z$ dimension in the reconstructed volume. This is indicated in the table as imaging system (2b).

Finally we implement (3) a confocal imaging system\cite{OToole2018} that images specific voxels $x_v$ of a volumetric space, illuminated with a focused ultrashort pulse of width $\sigma$. 
 
Note that in our case the \textit{virtual} imaging system is confocal, while the data for $H$ is not necessarily captured with a confocal arrangement. In the design of this system we can choose the phasor field pulse width $\sigma$. As this width increases, the depth resolution of the virtual imaging system worsens, although the signal-to-noise ratio improves. In practice, we found that a pulse width of about $\sigma = 6\,\omega$  yields the best results.
Ideally, we would apply this camera to an $H$ that is captured sampling both $x_p \in P$ and $x_c \in C$ with a set of points. In our current setup we only have a single SPAD detector and thus only obtain a single point for $x_c$. Using Helmholz Reciprocity we can interpret these datasets as having a single $x_p$ and and array of $x_c$. Improved results are anticipated when we capture a full $H$ which is left for future work as it is hard to implement without an array sensor (currently under development).

Following Helmholz reciprocity any linear optical system the path of light can be reversed without otherwise affecting the system. This is a manifestation of the time invariance of the wave equation. In practice this means that in a captured function $H(x_p \rightarrow x_c, t)$ the sets $x_p$ and $x_c$ are interchangeable. We make use of this property since our current imaging hardware scans the laser and only uses one stationary SPAD. We consider the array of laser positions to be our \textit{camera} points $x_c$ and the SPAD position the location of our light source $x_p$. The choice of capture arrangement is made for convenience since it is easier to calibrate the position of the laser spot on the wall. In future systems multiple SPADs will be used to eliminated the need for scanning or to provide more data points.
\begin{table}
\label{table:examples}
\centering
\begin{tabular}{ c|c|c }
  System & $\mathcal{P}(x_p,t)$ & $\Phi\left(\mathcal{P}(x_c,t)\right)$ \\
  \hline
  (1) Photo Camera & $e^{i\omega t}$ & $|\int_C{\mathcal{P}(x_c, t - \frac{1}{c}|x_v-x_c|)dx_c}|$ \footnotesize{($t$ is arbitrary)} \\
  (2a) Transient Camera & $e^{i\omega t}\delta(x_p-x_{ls})e^{i\frac{(t-t_0)^2}{2\sigma^2}}$  & $|\int_C{\mathcal{P}(x_c, t_f - \frac{1}{c}|x_v- x_c|)d x_c}|$  \\
  (2b) Corrected Transient Camera & $e^{i\omega t}\delta(x_p-x_{ls})e^{i\frac{(t-t_0)^2}{2\sigma^2}}$  & $[|\int_C{\mathcal{P}(x_c, t_f - \frac{1}{c}|x_v- x_c|)d x_c}|]_{max:z}$  \\
  (3) Confocal & $e^{i\omega t}e^{ik|x_v-x_p|}e^{i\frac{(t-t_0)^2}{2\sigma^2}}$ & $|\int_C{\mathcal{P}(x_c, - \frac{1}{c}|x_v-x_c|)dx_c}|$ \\
\end{tabular}
\caption{Functions needed to implement the three example cameras. The evaluation functions essentially describe the imaging transform of a lens with the resulting image being read out at different times with respect to the illumination. }
\end{table}

\subsection{Resolution Limits}
\label{supp:resolution}
Resolution limits for a NLOS imaging system with aperture diameter $d$ at imaging distance $L$ have been proposed\cite{Buttafava2015} to be the Rayleigh diffraction limit $\Delta x=1.22 c\tau d/L$, with $c$ the speed of light, for a pulse of full width at half maximum $\tau$. A derivation of a resolution limit is provided by O'Toole et al.\cite{OToole2018} (Equation 27 in the supplement), resulting in a similar formula of $\Delta x=0.5 c\tau d/L\approx 0.5 \lambda d/L$. However, this criterion for a resolvable object is based on the separability of the signal in the raw data and not in the reconstruction. Using our insight into the virtual camera, we can formulate a resolution limit that ensures a minimum contrast in the reconstruction. We also find that the details of the resolution limit depend of the particular choice of virtual projector and camera.

To estimate the resolution of a NLOS imaging system we use well-known methods that establish the resolution of wave based imaging system. For an imaging system that uses focusing only on the detection or illumination side, this limit is approximated by the Rayleigh criterion. For a virtual imaging system that provides focusing both on the light source and the detector side, the resolution doubles (as it does for example in a confocal or structured illumination microscope) and becomes $\Delta x=0.61 d\lambda/L$.

\subsection{Complexity analysis}
\label{supp:complexity}
We compare the complexity of our point-scanner system for 3D NLOS reconstructions, which is the most complex of our three example systems, against previous works. For this system the algorithm can be divided into two steps: A set of 1D convolutions, and a back-projection. 

Let us consider a dataset $H$ with $N_p$ discrete samples for $x_p\in P$, $N_c$ samples for $x_c \in C$, and $N_t$ temporal samples, and a 1D convolution kernel modeling the source phasor field $\mathcal{P}(x_s,t)$ of constant size $N_k$. 
The 1D convolutions $H(x_s \rightarrow x_c,t) \star \mathcal{P}(x_s,t)$ have a cost linear with respect to the size of the dataset $H$, and therefore $O(N_p N_c N_t N_k)=O(N_p N_a N_t)$ where the constant $N_k$ has been dropped.
Note that complexity in this case is not reduced when replacing the discrete Fourier transform with a fast Fourier transform due to the constant and relatively small size of the convolution kernel. 

The second operation is the back-projection. If we assume that the total number of samples in the reconstruction $N_v$ equals approximately the total number of samples in the data we can choose a voxel grid with side length $N_{x,y,z}$ such that $N_p N_a N_t\geq N_v=N_x N_y N_z$. We also assume that we have more samples along the time dimension than any of the other dimensions of $H$ ($N_t\geq N_p, N_a$). This is true for existing datasets. A naive backprojection\cite{Velten2012} has a complexity of $O(N_p N_c N_v)=O(N_p^2 N_c^2 N_t)$, which dominates over the convolution. 
Using the faster back-projection method described by Arellano et al.\cite{Arellano2017} reduces this complexity to $O(N_p N_a N_v^{1/3})\leq O(N_p N_a N_t)$ if $N_t\geq N_p, N_a$. In this case, the complexity of both convolution and backprojection is the same, leaving the complexity of our method as $O(N_p N_a N_t)$. We compare the complexity of our method against previous work in Table~\ref{tb:comparison} (we assume $N_p=N_a=N_t=N_x=N_y=N_z=N$).

\begin{table}
\centering
\begin{tabular}{c|c c c c c}
\hline
Method & FBP\cite{Velten2012} & CNLOS\cite{OToole2018} & Fast FBP\cite{Arellano2017} & Ours  \\
\hline
Complexity & $N^5$ & $N^3 \log(N^3)$ & $N^3$ & $N^3$   \\
\hline
\end{tabular}
\caption{Complexity of NLOS reconstruction methods: Filtered back-projection\cite{Velten2012} (FBP), confocal setup\cite{OToole2018} (CNLOS), fast filtered back-projection\cite{Arellano2017} (Fast FBP), and our method (Ours). 
}
\label{tb:comparison}
\end{table}

\subsection{Fresnel Approximation}
\label{supp:fresnel}
Using the RSD propagator we can formulate a point-to-point imaging operator like the ones shown in Table~\ref{table:examples}. Although this operator provides an exact solution of the wave equation, this propagator is rarely employed in optics due to its high computational cost and complexity. The Fourier optics framework provides multiple more efficient methods that we can utilize in our camera operators, from which we will employ the widely used Fresnel approximation.

If we constrain the operator to propagation between two parallel planes $\mathcal{S}$ and $\mathcal{D}$ separated by a distance $z$ we can write a plane-to-plane RSD operator with individual Cartesian components $(x_s,y_s) \in \mathcal{S}$ and  $(x_d,y_d) \in \mathcal{D}$ as
\begin{equation}
\mathcal{P}(x_d, y_d) =
	\frac{1}{i \lambda} \iint_S{\mathcal{P}(x_s,y_s) \frac{e^{ik\sqrt{(x_s-x_d)^2+(y_s-y_d)^2+z^2}}}{\sqrt{(x_s-x_d)^2+(y_s-y_d)^2+z^2}}dx_sdy_s}.
\end{equation}

We then make the assumption that $z >> (x_s-x_d)$ and $z >> (y_s-y_d)$, and since the factor $1/\sqrt{(x_s-x_d)^2+(y_s-y_d)^2+z^2}$ is a monotonous slowly varying function we can safely approximate it by $1/z$. The term in the argument, however, is much more sensitive since $e^{ik\sqrt{(x_s-x_d)^2+(y_s-y_d)^2+z^2}}$ is rapidly oscillating. To approximate it we perform a binomial expansion of the square root as
\begin{equation}
\sqrt{1+b}=1+\frac12 b-\frac18 b^2+...;
\end{equation}
Neglecting all but the first two terms of the expansion leads to
\begin{equation}
\begin{split}
\sqrt{(x_s-x_d)^2+(y_s-y_d)^2+z^2}
	&= z\sqrt{1+\left(\frac{x_s-x_d}{z}\right)^2+\left(\frac{y_s-y_d}{z}\right)^2}\\
	&\approx z\left(1+\left(\frac{x_s-x_d}{z}\right)^2+\left(\frac{y_s-y_d}{z}\right)^2\right).
\end{split}
\end{equation}

The RSD operator can thus be approximated to be
\begin{equation}
\begin{split}
\mathcal{P}(x_d, y_d)
	&\approx \frac{1}{i \lambda} \iint_S{\mathcal{P}(x_s,y_s) \frac{e^{ikz\left(1+\left(\frac{x_s-x_d}{z}\right)^2+\left(\frac{y_s-y_d}{z}\right)^2\right)}}{z}dx_sdy_s} \\
    &= \frac{1}{i \lambda} \frac{e^{ikz}}{z}\iint_S{\mathcal{P}(x_s,y_s)e^{ik\frac{\left(x_s-x_d\right)^2+\left(y_s-y_d\right)^2}{2z}}dx_sdy_s}.
\end{split}
\end{equation}
which can be interpreted as a 2D convolution
\begin{equation}
\mathcal{P}(x_d, y_d) \approx
	\mathcal{P}(x_s,y_s)\star \left(\frac{e^{ikz}}{i \lambda z}e^{ik\frac{x_s^2+y_s^2}{2z}}\right).
\end{equation}

We can use this approximation for the RSD propagator in all our camera operators. Reconstruction of a single focal plane can then be computed as a 2D convolution. The criteria for the validity of the Fresnel approximation is well known\cite{Born1999,Goodman2005} and given by
\begin{equation}
\frac{d^4}{4L^3\lambda}<<1
\end{equation}
where $d$ is the effective aperture radius of the virtual camera, $L$ is the distance between the aperture and the focal plane, and $\lambda$ is the wavelength.

\subsection{Comparison with other methods using the confocal dataset}
\label{supp:CNLOS_cmp}
In this section we show a comparison or different NLOS methods using the publicily available confocal dataset\cite{OToole2018}, which was captured using co-located transmission and detection spots on the relay wall ($x_p=x_c$). This set can be reconstructed using different NLOS methods, comparing their performance in simple scenes: the cone beam transform proposed by the original authors\cite{OToole2018}, filtered backprojection\cite{Velten2012, Gupta2012, Buttafava2015}, and our proposed virtual wave method. For this kind of confocal measurement datasets, backprojection steps can be expressed as a convolution with a pre-calculated kernel, and thus all three methods are using the same backprojection operator.

Note that neither our method nor filtered backprojection are limited to confocal data, and can be acquired making use of simpler devices and capture configurations. They can thus can be applied to a broader set of configurations and considerably more complex scenes.

For the CNLOS deconvolution method\cite{OToole2018}, we leave the optimal parameters unchanged. For our proposed virtual wave method, we use the aperture size and its spatial sampling grid published in the supplementary materials\cite{OToole2018} to calculate the optimal phasor field wavelength. For the filtered backprojection method it is important to choose a good discrete approximation of the Laplacian operator in the presence of noise. Previous works implicitly do the denoising step by adjusting the reconstruction grid size to approximately match the expected reconstruction quality\cite{Velten2012, Gupta2012, Buttafava2015}, or by downsampling across the measurements\cite{OToole2018}. All of them can be considered as proper regularizers. To provide a fair comparison without changing the reconstruction grid size, we convolve a Gaussian denoising kernel with the Laplacian kernel, resulting in a LOG filter, which we apply over the backprojected volume.

\begin{figure*}[ht!]
\centering
\resizebox{\textwidth}{!}{%
\includegraphics{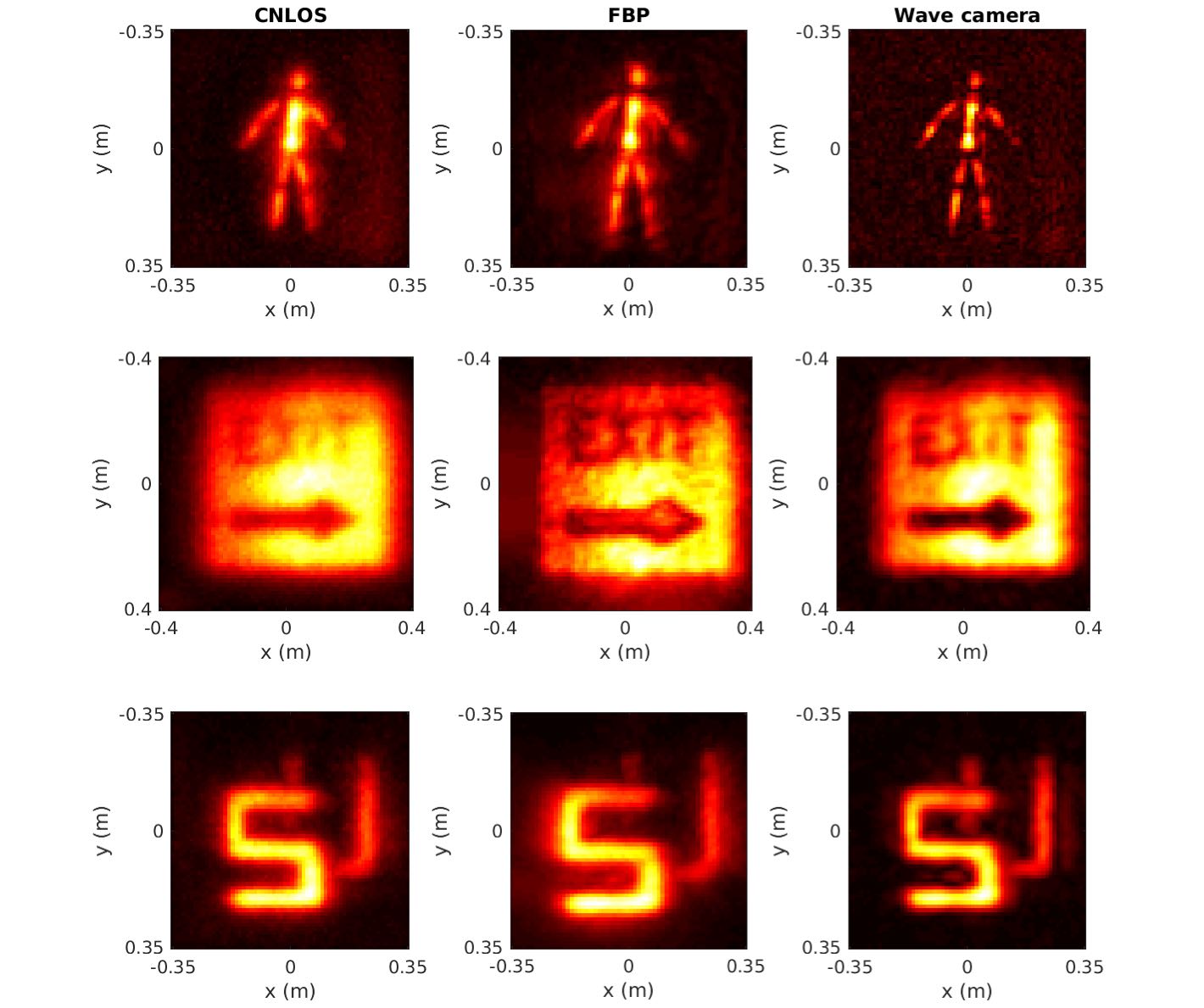}%
}
\caption{\textbf{Public data comparison}. Each row represents a different dataset. From left to right column, each represents CNLOS deconvolution, filter (LOG) backprojection, and our proposed virtual wave imaging method. All methods use the same datasets without any preprocessing and visualize in the same way. 
}
\label{fig:Supp_CNLOS_exp_comparison_2}
\end{figure*}

From this dataset, which contains isolated objects of simple geometries, all methods deliver reconstructions approaching their resolution limits,  with improved contrast and cleaner contours in our wave camera method (see Figure~\ref{fig:Supp_CNLOS_exp_comparison_2}). 
The improved contrast in our method is due to better handling multiply scattered light, which polutes the reconstructions in the other methods. 
While this scattering can be ameliorated in simple datasets by thresholding, this strategy fails in more complex scenes like the ones shown in the main paper.

\begin{figure*}[ht!]
\centering
\resizebox{\textwidth}{!}{%
\includegraphics{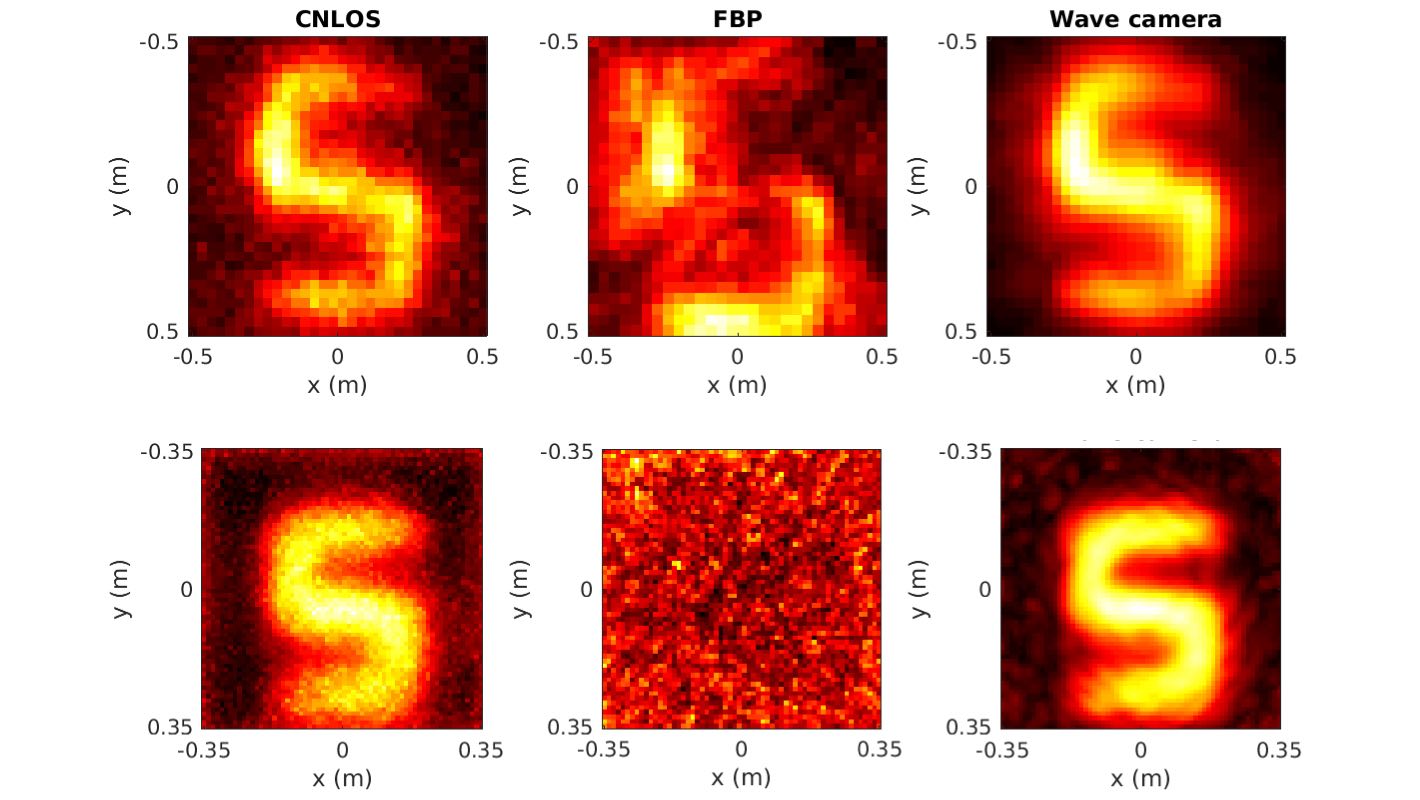}%
}
\caption{\textbf{Public data comparison (cont.)}. Each row represents a different dataset. From left to right column, each represents CNLOS deconvolution, filter (LOG) backprojection, and our proposed virtual wave imaging method. All methods use the same datasets without any preprocessing and visualize in the same way. 
}
\label{fig:Supp_CNLOS_exp_comparison_3}
\end{figure*}

In the noisy datasets (Figure~\ref{fig:Supp_CNLOS_exp_comparison_3}), filtered backprojection fails. CNLOS includes a Wiener filter that performs well at removing uniform background noise, although a noise level must be explicitly estimated. Our phasor-field virtual wave method, on the other hand, performs well automatically, without the need to explicitly estimate a noise level. This is particular important in complex scenes with MPI where background is rarely uniform across the scene.

\end{document}